\documentclass[runningheads]{llncs}
\usepackage{graphicx}
\usepackage{amsmath,amssymb} 
\begin{document}
\title{Thinking Hallucination for Video Captioning}
%
%
\author{Nasib Ullah\inst{1}\thanks{This work was done while Nasib Ullah was a PLP at ECSU, ISI Kolkata, India.\\Pre-print version } \and
Partha Pratim Mohanta\inst{2} }

%
\authorrunning{N. Ullah et al.}
%
\institute{LIVIA, Dept. of Systems Engineering, ÉTS, Montreal, Canada \\
\email{nasibullah.nasibullah.1@ens.etsmtl.ca}\\
 \and
ECSU, Indian Statistical Institute, Kolkata, India\\
\email{ppmohanta@isical.ac.in}}
\maketitle              
\begin{abstract}
With the advent of rich visual representations and pre-trained language models, video captioning has seen continuous improvement over time. Despite the performance improvement, video captioning models are prone to hallucination. Hallucination refers to the generation of highly pathological descriptions that are detached from the source material. In video captioning, there are two kinds of hallucination: object and action hallucination. Instead of endeavoring to learn better representations of a video, in this work, we investigate the fundamental sources of the hallucination problem. We identify three main factors: (i) inadequate visual features extracted from pre-trained models, (ii) improper influences of source and target contexts during multi-modal fusion, and (iii) exposure bias in the training strategy. To alleviate these problems, we propose two robust solutions: (a) the introduction of auxiliary heads trained in multi-label settings on top of the extracted visual features and (b) the addition of context gates, which dynamically select the features during fusion. The standard evaluation metrics for video captioning measures similarity with ground truth captions and do not adequately capture object and action relevance. To this end, we propose a new metric, COAHA (caption object and action hallucination assessment), which assesses the degree of hallucination. Our method achieves state-of-the-art performance on the MSR-Video to Text (MSR-VTT) and the Microsoft Research Video Description Corpus (MSVD) datasets, especially by a massive margin in CIDEr score.

\end{abstract}
\section{Introduction}

Video captioning is the translation of a video into a natural language description. It has many potential applications, including video retrieval, human-computer interface, assisting visually challenged, and many more. The encoder-decoder based sequence to sequence architecture (initially proposed for machine translation) has helped video captioning exceptionally. Pre-trained vision models extract the frame features at the encoder section, and the decoder is a conditional language model. Recent improvements happened in broadly three areas, better visual feature extraction \cite{marn:1,stg_kd:1}, better and support of external pre-trained language models \cite{hrne:1,org_trl:1} and, better strategy \cite{picknet:1,sgn:1} to sample informative frames.

\begin{figure}[t]
\begin{center}
    \includegraphics[width=0.7\textwidth]{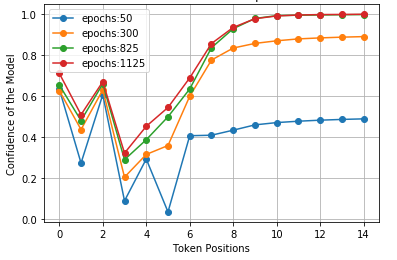}
    \caption{The plot of the model confidence at the different token positions. The plot is done by averaging confidence over validation set on MSVD data and using the model MARN \cite{marn:1}.}
    \label{fig:confidence_intro}
    \end{center}
\end{figure}

Despite the improvements, a significant problem remains that is known as a hallucination \cite{hal_ic:1,hal_nmt:1,hal_nmt:6} in the literature.
\begin{figure}[t]
\begin{center}
    \includegraphics[width=0.7\textwidth]{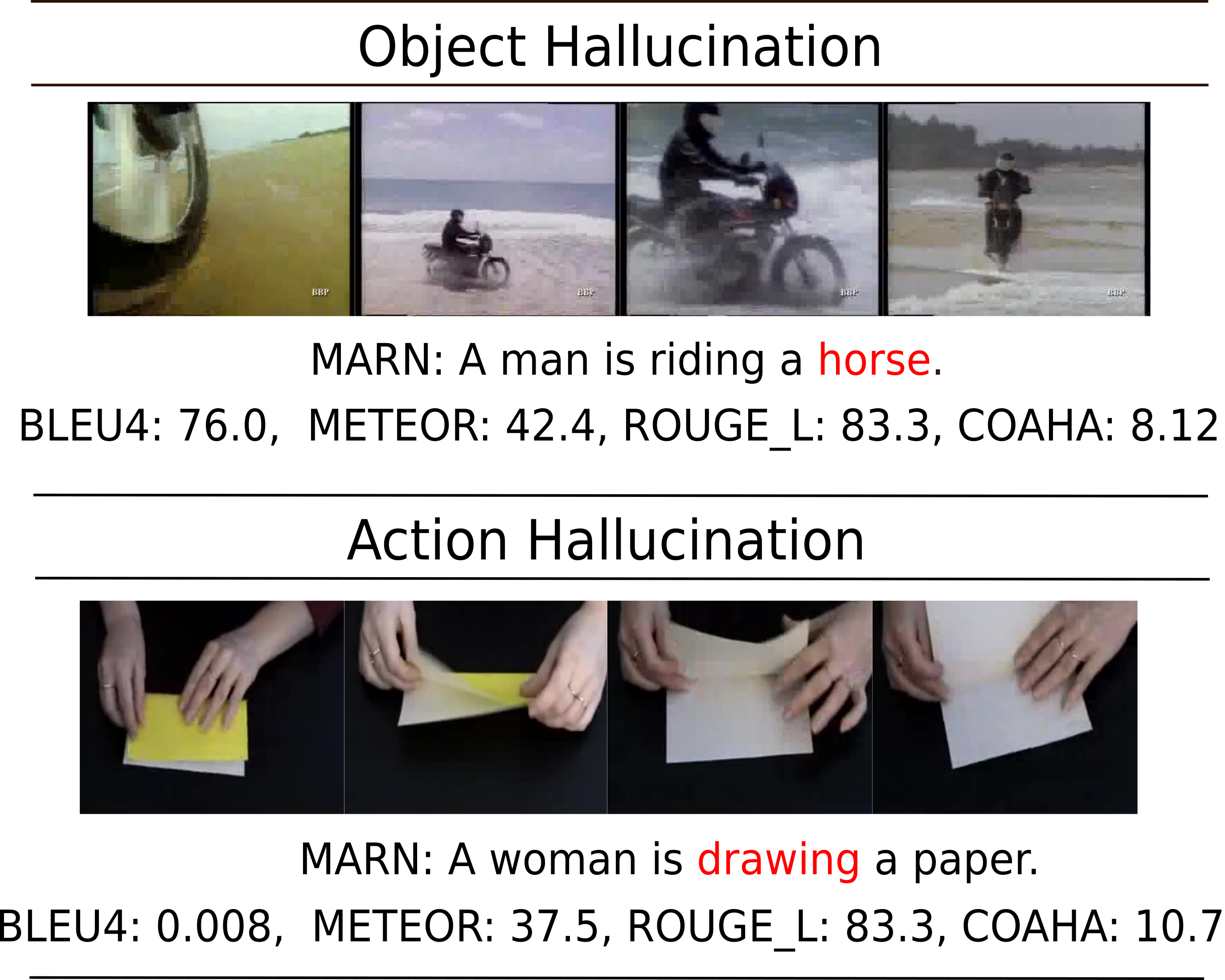}
    \caption{Video Captioning models suffer from two types of hallucination, Object hallucination (above) and action hallucination (below). Unfortunately, the standard metrics fail to adequately capture hallucination, whereas our proposed COAHA metric is more relevant to hallucination. The results are generated using the MARN \cite{marn:1} model. For COAHA, lower is better. }
    \label{fig:hal_intro}
\end{center}
\end{figure}
Unlike image captioning, there are two types of hallucination in the case of video captioning: object hallucination occurs when the model describes objects that are not present in the video, and action hallucination occurs when the model describes an action that has not been performed in the video as depicted in Fig.~\ref{fig:hal_intro}. Another evidence to understand that hallucination is a significant problem can be seen from the token position vs. model confidence plot in Fig.~\ref{fig:confidence_intro}. Generally, the object and action words occur towards the middle of the caption. Thus, we can see that the model has low confidence while generating tokens related to objects and actions.

In this work, we point out three fundamental sources of hallucination for video captioning: (i) inadequate visual features from pre-trained models, (ii) improper influence of features during intra-modal and multi-modal fusion, and (iii) exposure bias in the model training strategy. We propose two robust and straightforward techniques that alleviate the problem. The exposure bias problem acts mainly under domain shift \cite{hal_nmt:1,hal_nmt:2}, so we will not be focusing on that. Standard evaluation metrics only measure similarity with ground truth caption and may not fully capture the object and action relevance. To this end, we propose a new metric COAHA (caption object and action hallucination assessment) to assess the hallucination rate. The CHAIR metric \cite{hal_ic:1} has been proposed before for image captioning. However, it considers only object hallucination and relies on object segmentation annotation unavailable for video captioning datasets. Unlike CHAIR, our metric considers both object and action hallucination and leverage existing captions instead of relying on segmentation annotations.

Finally, we show that our solutions outperform the state-of-the-art models in standard metrics, especially by a large margin in CIDEr-D on Microsoft Research Video Description Corpus (MSVD) and MSR-Video to Text (MSR-VTT) datasets.

\section{Related Work}
\label{sec:related_work}
\subsection{Video Captioning.} 
The main breakthrough in video captioning happened with the encoder-decoder based sequence to sequence architecture. Although MP-LSTM \cite{mp_lstm:1} applied this first, they did not capture the temporal information between frames. S2VT \cite{s2vt:1} and SA-LSTM \cite{sa_lstm:1} capture the temporal dependencies between frames. The former shares a single LSTM network for both encoder and decoder, whereas the latter uses attention weights over frames and 3D HOG features.  Recent methods have improved upon  SA-LSTM \cite{sa_lstm:1}. For example, RecNet \cite{recnet:1} has added reconstruction loss based on backward flow from decoder output to the video feature construction. MARN \cite{marn:1} has used external memory to capture various visual contexts corresponding to each word. Both MARN \cite{marn:1} and M3 \cite{m3:1} utilize motion and appearance features. However, unlike MARN \cite{marn:1}, M3 \cite{m3:1} uses heterogeneous memory to model long-term visual-text dependency. OA-BTG \cite{oa_btg:1} and STG-KD \cite{stg_kd:1} both use object features along with appearance and motion features. OA-BTG \cite{oa_btg:1} leverage the temporal trajectory features of salient objects in the video. In contrast, STG-KD \cite{stg_kd:1} uses a spatio-temporal graph network for object interaction features and a Transformer network for the language model instead of recurrent neural networks. ORG-TRL \cite{org_trl:1} also uses object interaction features, but unlike STG-KD \cite{stg_kd:1}, it utilizes an external language model which guides the standard recurrent neural network based language model. Another line of work focuses on a better sampling strategy to choose informative video frames. PickNet \cite{picknet:1} samples informative frames based on reward-based objectives, whereas SGN \cite{sgn:1} uses partially decoded information to sample frames. Regardless of the improvements, hallucination is still one of the major problems in existing models. In this work, we will be focusing on significant sources of hallucination, how to mitigate them, and finally, a metric to measure the rate of hallucination.

\subsection{Hallucination Problems in Natural Language Generation.}
Koehn et al. \cite{hal_nmt:3} pointed out the problem of hallucination among the six challenges in neural machine translation. M{\"{u}}ller et al. \cite{hal_nmt:2} and Wang et al. \cite{hal_nmt:1} linked the problem of hallucination with exposure bias. Wang et al. \cite{hal_nmt:1} have used minimum risk training (MRT) instead of Maximum likelihood (MLE) to reduce the exposure bias and concluded the improvement mostly happens under domain shift. Lee et al. \cite{hal_nmt:5} has claimed that NMT models hallucinate under source perturbation. Tu et al. \cite{hal_nmt:4} has used the context gates to control the source and target contribution, reducing the hallucination for neural machine translation. In the case of image captioning, Anna et al.  \cite{hal_ic:1} discussed object hallucination and proposed CHAIR metric to assess the degree of object hallucination. Unlike neural machine translation and image captioning, video captioning is more complex because of the spatio-temporal nature of the input data. Although state-of-art video captioning models suffer from hallucination, there is no pre-attempt. To the best of our knowledge, this is the first attempt to address hallucination for video captioning.    
\begin{figure*}
    \begin{center}
    \includegraphics[width=\textwidth]{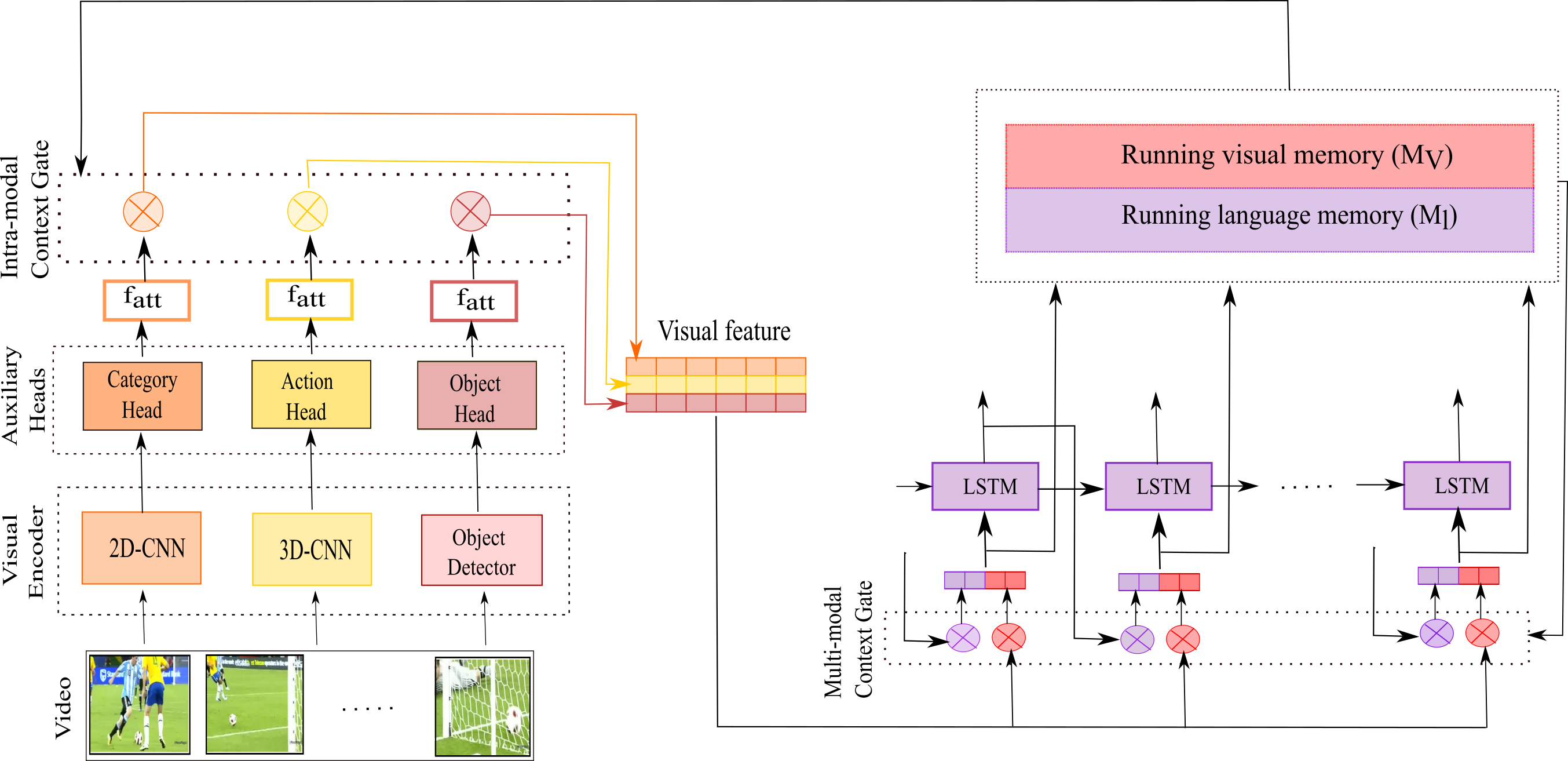}
    \caption{The architecture of our proposed model. The components are (1) Visual Encoder for visual feature extraction, (2) auxiliary heads, and (3) context gates to reduce hallucination, and finally, (4) Decoder for sentence generation. }
    \label{fig:model_arc}
    \end{center}
\end{figure*}

\section{Methodology}
As shown in Fig.~\ref{fig:model_arc}, the components of our proposed system are (a) Visual Encoder responsible for the extraction of different visual features from the input video, (b) Auxiliary heads to reduce the hallucination that occurs from inadequate visual features, (c) Context gates to dynamically control the importance of features during intra-modal and multi-modal fusion and, (d) Decoder which is a conditional language model. We utilize running visual and language memory along with these components to act as an external short-term memory. We have used multi-label Auxiliary head loss and Coherent loss along with the standard cross-entropy loss for training.

\subsection{Visual Encoder}
From a given input video, we uniformly sample N frames $\{f_{i}\}_{i=1}^{N}$ and clips $\{c_{i}\}_{i=1}^N$, where each $c_{i}$ is a sequence of frames around $f_{i}$. We extract appearance features $\{a_{i}\}_{i=1}^N$ and motion features $\{m_{i}\}_{i=1}^N$ using pre-trained 2D CNN $\phi^a$ \cite{vit:1} and 3D CNN $\phi^m$ \cite{c3d:1} respectively, where $a_{i} = \phi^a(f_{i})$ and $m_{i} = \phi^m(c_{i})$. Apart from appearance and motion, we also extract object features $\{o_{i}\}_{i=1}^N$
using a pre-trained object detection module $\phi^o$ \cite{fasterrcnn:1} where $o_{i} = \phi^o(f_{i})$. For each frame, we extract salient objects based on the objectiveness threshold $v$ and average the features of those salient objects. Appearance ($\{a_{i}\}_{i=1}^N$) and motion ($\{m_{i}\}_{i=1}^N$) feature helps to understand the global context and the motion information of the video. In contrast, the object features ($\{o_{i}\}_{i=1}^N$) are more localized and helps to understand fine-grained information. 

\subsection{Auxiliary Heads: Category, Action, and Object}
\label{subsection:aux_head}
Unlike image captioning, the sequence of frames in video captioning does make the training of the feature extraction module computationally extensive. Existing methods use pre-trained models for different visual feature extraction and consider these features as the input to the model. However, the pre-trained models are trained on different data distribution, and extracted features are not adequate, which leads to the hallucination problem. We hypothesize that inadequate object and appearance features are responsible for object hallucination, whereas inadequate motion features lead to action hallucination.

\begin{figure}[t]
\begin{center}
    \includegraphics[width=0.7\textwidth,height=150px]{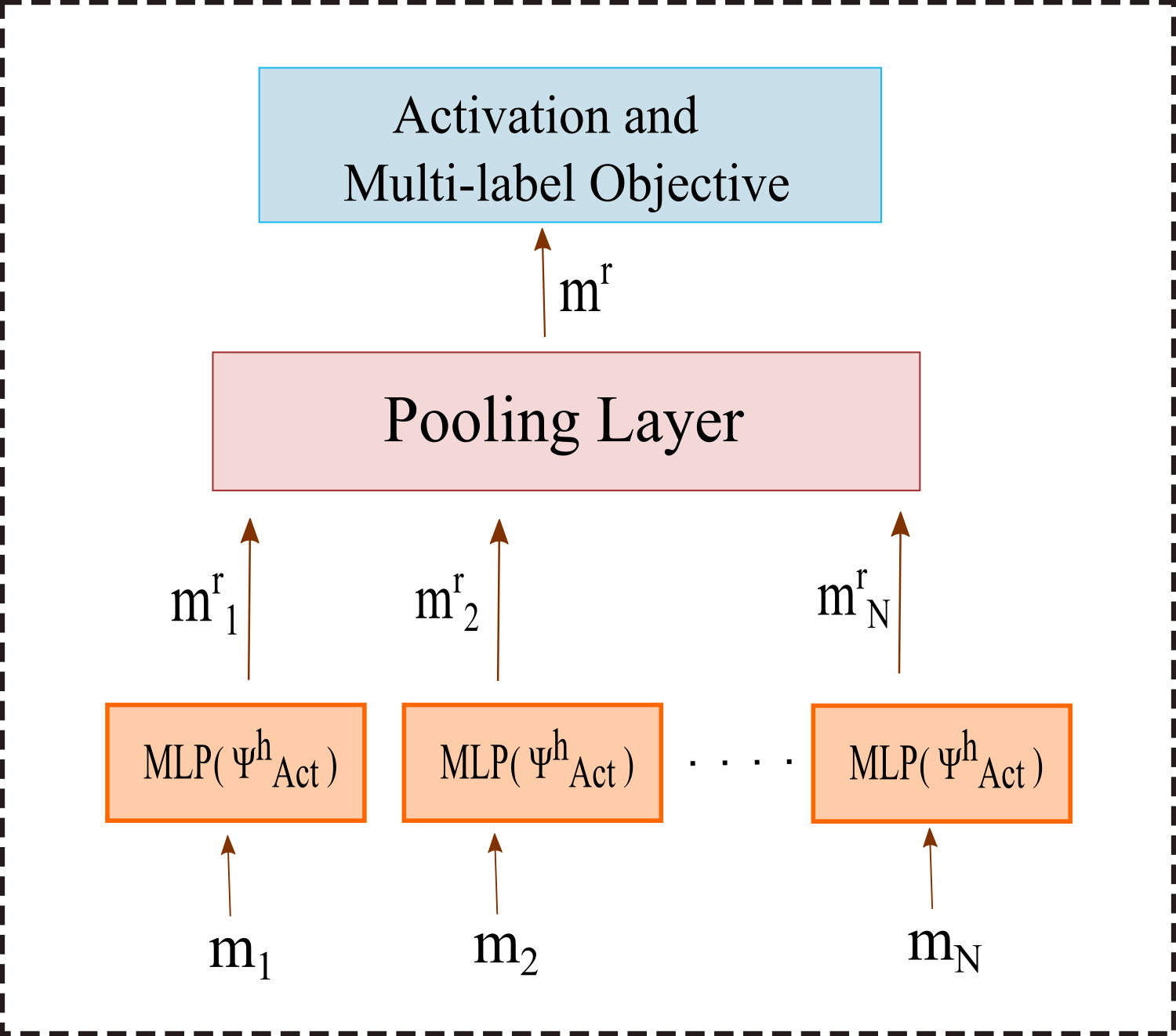}
    \caption{Diagram of action auxiliary head. $\{m_{i}\}_{i=1}^N$ are the motion features from visual encoder. The MLP is shared over clip features. The same architecture follows for object and categorical auxiliary heads.}
    \label{fig:action_head}
    \end{center}
\end{figure}

We propose specialized classification heads on top of each type of visual feature, and they are trained in multi-label settings. The supervision is generated by leveraging existing captions, and the features from the heads are more suitable and semantically relevant. We include object head $\psi_{Obj}^h$ on top of the object features $\{o_{i}\}_{i=1}^N$ , where the training objective is to predict visual objects present in the caption for that input video. Similarly, action head $\psi_{Act}^h$ applies to motion features $\{m_{i}\}_{i=1}^N$ to predict actions present in the caption. The ground truth for both the action head and object head are extracted using Named Entity Recognition \cite{ner:1,dep:1} and Parts-of-speech tagging \cite{pos_cg:1} methods from NLP literature. To refine the appearance features $\{a_{i}\}_{i=1}^N$, we add categorical head $\psi_{Cat}^h$ to predict the video category, which is available for the MSR-VTT dataset but not for MSVD. Although we have used three specific objectives, this framework can be applied with other objectives, such as predicting attributes on top of the appearance feature. Finally, we use the features $\{a_{i}^r\}_{i=1}^N$, $\{m_{i}^r\}_{i=1}^N$ and $\{o_{i}^r\}_{i=1}^N$ from the specific auxiliary head as the input to the model, where $ a_{i}^r = \psi_{Cat}^h(a_{i})$,   $m_{i}^r = \psi_{Act}^h(m_{i})$  and  $o_{i}^r = \psi_{Obj}^h(o_{i})$. The Fig.~\ref{fig:action_head} shows the modeling of the action head.

\subsection{Context Gates: Balancing Intra-Modal and Multi-Modal Fusion }

There are two types of fusion of features in the existing video captioning framework: intra-modal fusion and inter-modal or multi-modal fusion. In intra-modal fusion, different visual (appearance, motion, and object) features are merged, and thus one feature might dominate other vital features. For example, motion features might dominate appearance and object features, which lead to object hallucination, and action hallucination occurs due to dominance of object and appearance features over motion features. Similarly, in the case of multi-modal fusion at the decoder, the dominant language prior leads to hallucination \cite{hal_nmt:7,hal_nmt:4}, whereas the dominant source context leads to a not fluent caption.

We have included two separate context gate units for intra-modal and multi-modal fusion of features, which mitigate the hallucination. The gates are responsible for dynamically selecting the importance of each type of feature. Tu et al. \cite{hal_nmt:4} have used the context gate before to balance the source and target context in neural machine translation. Nevertheless, unlike Tu et al. \cite{hal_nmt:4}, we have used separate gates for intra-modal and multi-modal fusion. Also, instead of depending on the decoder memory, our gates are conditioned on running visual ($M_{v}$) and language ($M_{l}$) memory as shown in Fig.~\ref{fig:model_arc}. Empirical studies have shown better performance with separate running memories rather than decoder memory. One possible reason might be the decoder memory is contaminated by a mixture of source and target features and noisy LSTM gates. 

\noindent At the encoder, the combined visual feature $v_{m,t}$ is,
\begin{equation}
v_{m,t} = [CG_{a}^E \circ a^r_{t}  ; CG_{m}^E \circ m^r_{t} ; CG_{o}^E \circ o^r_{t}] 
\end{equation}
\begin{center}
where,    
\end{center}
\begin{equation}
 a^r_{t} = \sum_{i=1}^N \alpha_{i,t} a_{i,t}^r,  m^r_{t} = \sum_{i=1}^N \alpha_{i,t} m_{i,t}^r, \\
 o^r_{t} = \sum_{i=1}^N \alpha_{i,t} o_{i,t}^r    
\end{equation}
\begin{equation}
   CG_{a/m/o}^E = f(a^r_{t},m^r_{t},o^r_{t},M_{v},M_{l})
\end{equation}
$CG_{a/m/o}^E$ are the context gates corresponding to appearance, motion, and object features, respectively, and $\alpha_{i,t}$ are the attention weights \cite{sa_lstm:1}. We share the attention network over the object, motion, and appearance features for the regularization purpose.

At the decoder, the multi-modal feature $C_{t}$ is,
\begin{equation}
\label{eq:combine_context}
C_{t} = [ CG_{S}^D \circ v_{m,t} ; CG_{T}^D \circ E(y_{t-1})  ]
\end{equation}
\begin{center}
where,
\end{center}
\begin{equation}
 CG_{S/T}^D = f(v_{m,t},E(y_{t-1}),M_{v},M_{l})
\end{equation}
$CG_{S/T}^D$ are the context gates corresponding to source and target context, respectively, and $E(y_{t-1})$ is the embedding feature of the previous token. $f$ is realized using MLP, [;] denotes concatenation, and $\circ$ denotes Hadamard product.
\subsection{Decoder}
The decoder is a language model conditioned on merged visual and language prior features $C_{t}$. We have used the LSTM network because of its superior sequential modeling performance. At $t$ time steps during the sentence generation, the hidden memory of LSTM can be expressed as,
\begin{equation}
    h_{t} = LSTM(C_{t},h_{t-1})
\end{equation}
where $h_{t-1}$ is the hidden memory at previous time steps, and $C_{t}$ is the combined multi-modal feature as shown in Equation \ref{eq:combine_context}.
Although transformer-based \cite{transformer:1} models have outperformed LSTM in most NLP benchmarks, LSTM and transformer models performed almost equally in our case. Finally, the probability distribution over words is generated using a fully connected layer followed by a softmax layer.
\begin{equation}
    P(y_{t}|V,y_{1},y_{2},..,y_{t-1}) = softmax(V_{h}h_{t}+b_{h})
\end{equation}
where $V_{h}$ and $b_{h}$ are learnable parameters.
\subsection{Parameter Learning}
Along with the standard cross-entropy loss, our model is trained using two additional losses, Auxiliary head loss, and Coherent loss.

\subsubsection{Attention Based Decoder.}
Negative log-likelihood (or cross-entropy) is the standard objective function for the video captioning models. The loss for a mini-batch is
\begin{equation}
    \textit{L}_{CE} = - \sum_{i=1}^B \sum_{t=1}^T \log p(y_{t}|V,y_{1},y_{2},..,y_{t-1};\theta)
\end{equation}
Where $\theta$ is learnable parameters, $V$ is the video feature, $y_{t}$ is the $t^{th}$ word in the sentence of length T, and B is the mini-batch size. 
\subsubsection{Auxiliary Heads.}
Auxiliary heads are trained in multi-label settings. For all three heads, we have used binary cross-entropy loss over sigmoid activation. The total auxiliary loss is,
\begin{equation}
    \textit{L}_{AH} = \lambda_{C} \textit{L}_{AH}^C + \lambda_{A} \textit{L}_{AH}^A + \lambda_{O} \textit{L}_{AH}^O
\end{equation}
where $\lambda_{C}$, $\lambda_{A}$ and $\lambda_{O}$ are hyperparameters corresponding to Category head loss $\textit{L}_{AH}^C$, Action head loss $\textit{L}_{AH}^A$, and Object head loss $\textit{L}_{AH}^C$, respectively. Each loss is calculated as
\begin{equation*}
\textit{L}_{AH}^C = \Upsilon (Y^C,\psi_{Cat}^h({a_{j}}_{j=1}^N)), \textit{L}_{AH}^A = \Upsilon (Y^A,\psi_{Act}^h({m_{j}}_{j=1}^N) )
\end{equation*}
\begin{equation*}
    \textit{L}_{AH}^O = \Upsilon (Y^O,\psi_{Obj}^h({o_{j}}_{j=1}^N)) 
\end{equation*}
\begin{center}
    where,
\end{center}
\begin{equation}
\Upsilon(y,p) = \sum_{i=1}^B \sum_{m=1}^M [ y_{i,m} \log p_{i,m} + (1 - y_{i,m}) \log (1-p_{i,m})]
\end{equation}
$Y^C$, $Y^A$ and $Y^O$ are category, action and object ground truth for the given batch ($B$) of data.
\subsubsection{Coherent Loss.}
The consecutive frames in a video are highly repetitive. So encoding of consecutive frames should be similar. We apply the coherent loss to restrict the embedding of consecutive frames to be similar. Coherent loss \cite{marn:1} has been used before to regularize attention weights, but unlike Pei et al. \cite{marn:1}, we apply coherent loss on appearance, motion, and object features. It is debatable whether to apply coherent loss on motion features, but empirically we got a good result, and the possible reason might be shorter clip length. The total coherent loss for a mini-batch is,
\begin{equation}
    \textit{L}_{CL} = \lambda_{fcl} \textit{L}_{CL}^a + \lambda_{mcl} \textit{L}_{CL}^m + \lambda_{ocl} \textit{L}_{CL}^o + \lambda_{acl} \textit{L}_{CL}^{\alpha}  
\end{equation}
where $\lambda_{fcl}$, $\lambda_{mcl}$, $\lambda_{ocl}$ and $\lambda_{acl}$ are hyperparameters corresponding to appearance coherent loss $\textit{L}_{CL}^a$, motion coherent loss $\textit{L}_{CL}^m$, object coherent loss $\textit{L}_{CL}^o$ and attention coherent loss $\textit{L}_{CL}^\alpha$ respectively.

\noindent The individual coherent losses are calculated as, $\textit{L}_{CL}^a=\Phi(a_{i}^r)$,     $\textit{L}_{CL}^m =\Phi(m_{i}^r)$, $\textit{L}_{CL}^o=\Phi(o_{i}^r)$ and           $\textit{L}_{CL}^{\alpha}=\Phi(\alpha_{i})$ where,
\begin{equation}
    \Phi(f) = \sum_{i=1}^B \sum_{t=1}^T \sum_{n=2}^N | f_{n,t}^{(i)} - f_{n-1,t}^{(i)} |
\end{equation}

\noindent The overall loss function of our model is
\begin{equation}
    \textit{L} = \textit{L}_{CE} + \textit{L}_{AH} + \textit{L}_{CL}
\end{equation}

\subsection{The COAHA Metric} In order to measure the degree of hallucination, we propose COAHA (caption object and action hallucination assessment) metric. The metric is defined for per instance and is given by,
\begin{equation}
    COAHA = OH+AH 
\end{equation}
\begin{center}
where,
\end{center}
\begin{equation}
    OH = \frac{\sum_{h_{i}\in H_{O}}d_{i}^h}{T}, \quad  AH = \frac{\sum_{a_{i}\in H_{A}}d_{i}^a}{T}
\end{equation}
$OH$ measures the degree of object hallucination, and $AH$ measures the degree of action hallucination. $T$ is the average ground truth caption length and $N_{O}$ and $N_{A}$ are the set of objects and actions mentioned in the video. $H_{O}$ is the set of hallucinated objects, and $H_{A}$ is the set of hallucinated actions. $d_{i}^h$ measures the average semantic distance between the current hallucinated object and all objects in $N_{O}$. Similarly, $d_{i}^a$ measures the average semantic dissimilarity of hallucinated action. The higher $d_{i}^h$ or higher cardinality of $H_{O}$ will lead to a higher value of $OH$, and the same follows for $AH$. So, if the hallucinated object or action is a synonym, then the semantic distance would be small, which will lead to a low hallucination value. Finally, we have taken the addition of $OH$ and $AH$. We have applied the same approach mentioned in the Auxiliary Head section to extract the list of objects and actions from each video (from the ground truth caption). For measuring the semantic distance ($d_{i}^h$ and $d_{i}^a$) we have used pre-trained Fasttext \cite{fasttext:1} embedding model. Mathematically,
\begin{equation}
    d_{i}^{h} = \frac{1}{|N_{O}|} \sum_{w_{k}\in N_{O}} \vartheta(w_{k},h_{i})
\end{equation}
where, $\vartheta$ is the cosine similarity on top of the Fasttext \cite{fasttext:1} embedding. Similarly the formulae follows for $d_{i}^a$.

\section{Experimental Results}
We evaluate our proposed model on two most popular benchmark datasets: MSVD and MSR-VTT. To compare with the state-of-the-art results, we have used the four most popular metrics: CIDEr \cite{cider:1}, METEOR \cite{meteor:1}, ROUGE-L \cite{rouge:1} and BLEU-4 \cite{bleu:1}.
\subsection{Datasets}
\textbf{MSVD.} Microsoft Video Description (MSVD) dataset \cite{msvd:1} consists of 1970 open domain videos of single activity and is described by 40 captions generated by Amazon Mechanical Turk. For a fair comparison, we have followed the standard split \cite{mp_lstm:1,sa_lstm:1} of 1200 videos for training, 100 for validation, and 670 for testing.\\
\textbf{MSR-VTT.} MSR Video-to-Text (MSR-VTT) dataset \cite{msrvtt:1} is a large scale benchmark datasets with 10k videos and 20 categories. Each video is annotated with 20 captions with an average length of 20 seconds. We have followed the standard benchmark split \cite{msrvtt:1} of 6513 for training, 497 for validation, and 2990 for testing.
\begin{table*}
    \centering
    \resizebox{0.90\textwidth}{!}{
    \begin{tabular}{c|c c c c|c c c c}
      \hline \hline 
     Models &   & MSVD &  &  &  & MSR-VTT &  &  \\
     & B@4  & M & R & C & B@4 & M & R & C \\ 
     \hline 
     SA-LSTM \cite{sa_lstm:1} & 45.3 & 31.9 & 64.2 &76.2 & 36.3 & 25.5 & 58.3 & 39.9 \\
     h-RNN \cite{hrnn:1} & 44.3 & 31.1 & - & 62.1 & - & - & - & - \\
     hLSTMat \cite{hlstmatt:1} &53.0 & 33.6 & - & 73.8 & 38.3 & 26.3 & - & - \\ 
     RecNet \cite{recnet:1} & 52.3 & 34.1 & 69.8 & 80.3 & 39.1 &26.6 & 59.3 &42.7 \\
     M3 \cite{m3:1} & 52.8 & 33.3 & - & - & 38.1 & 26.6 & - & - \\
     PickNet \cite{picknet:1} & 52.3 & 33.3 & 69.6 & 76.5 & 41.3 & 27.7 & 59.8 & 44.1 \\
     MARN \cite{marn:1} & 48.6 & 35.1 & 71.9 & 92.2 & 40.4 & 28.1 & 60.7 & 47.1 \\
     GRU-EVE \cite{gru_eve:1} & 47.9 & 35.0 & 71.5 & 78.1 & 38.3 & 28.4 & 60.7 & 48.1 \\
     POS+CG \cite{pos_cg:1} & 52.5 & 34.1 & 71.3 & 88.7 & 42.0 & 28.2 & 61.6 & 48.7 \\
     OA-BTG \cite{oa_btg:1} & \textbf{56.9} & 36.2 & - & 90.6 & 41.4 & 28.2 & - & 46.9 \\
     STG-KD \cite{stg_kd:1} & 52.2 & \textbf{36.9} & 73.9 & 93.0 & 40.5 & 28.3 & 60.9 & 47.1 \\ 
     SAAT \cite{saat:1} & 46.5 & 33.5 & 69.4 & 81.0 & 40.5 & 28.2 & 60.9 & 49.1 \\
     ORG-TRL \cite{org_trl:1} & 54.3 & 36.4 & 73.9 & 95.2 & \textbf{43.6} & 28.8 & \textbf{62.1} & 50.9 \\
     SGN \cite{sgn:1} & 52.8 & 35.5 & 72.9 & 94.3 & 40.8 & 28.3 & 60.8 & 49.5 \\
     \hline  
     Ours & 53.3 & 36.5 & \textbf{74.0} & \textbf{99.9} & 41.1 & \textbf{28.9} & 61.9 & \textbf{51.7} \\
     \hline \hline
    \end{tabular}}
    \caption{Performance comparison on MSVD and MSR-VTT benchmarks. B4, M, R, and C denote BLEU{-}4, METEOR, ROUGE\_L, and CIDEr, respectively.}
    \label{tab:main_table}
\end{table*}
\subsection{Implementation Details}
\textbf{Feature Extraction and Decoding.}We uniformly sample 28 frames per video, and sentences longer than 30 words are truncated. The 1024-D appearance features of each frame are extracted by ViTL \cite{vit:1} pre-trained on Imagenet \cite{imagenet:1}. For the 2048-D motion features, we use C3D \cite{c3d:1} with ResNeXt-101 \cite{resnext:1} and pre-trained on the Kinetic-400 dataset. To extract the object features, we apply Faster-RCNN \cite{fasterrcnn:1} with ResNet-101 and FPN backbone pre-trained on MSCOCO \cite{detectron2:1}. Appearance, motion, and objects features are embedded to 512-D before sending to the next stage. At the decoder LSTM, we use a hidden layer size of 512, and the word embedding size is fixed to 512.The vocabulary is built by words with at least 5 occurrences. During testing, we use greedy decoding to generate the sentences.\\
\textbf{Auxiliary Heads and Context Gates Related Details.} The running visual and language memory size is set to 512. We use the Spacy \cite{spacy:1} module pre-trained with transformers \cite{transformer:1} for object and action extraction from ground truth captions. We have also used Porter Stemmer \cite{nltk:1} to work with the root form of words. Finally, to measure the semantic distance between words, we have used the pre-trained Fasttext \cite{fasttext:1} model from Gensim \cite{gensim:1}.\\
\textbf{Other Details.} We apply Adam \cite{adam:1} with a fixed learning rate of 1e-4 and a gradient clip value of 5. Our model is trained for 900 epochs with a batch size of 100. The coherent loss weights $\lambda_{acl}$, $\lambda_{fcl}$, $\lambda_{mcl}$, and $\lambda_{ocl}$ are set as 0.01, 0.1, 0.01, and 0.1, respectively. All the experiments are done in a single Titan X GPU.
\subsection{Quantitative Results}
To proclaim the effectiveness of our approach, we compare the performance of our model with state-of-the-art models. 
\begin{table}[h]
\centering
    \resizebox{0.7\columnwidth}{!}{
    \begin{tabular}{c c |c c c c| c}
    \hline \hline
        Methods &  &  &  & & &  \\
        A.Heads & C.Gates & B@4 & M & R & C & COAHA \\
    \hline
     $\times$ & $\times$ & 50.1 & 34.8 & 72.9 & 91.1 & 10.57 \\
     \checkmark & $\times$ & 52.6 & 36.2 & 73.4 & 96.6 & 7.90 \\
     $\times$ & \checkmark & 52.5 & 36.3 & 73.6 & 97.1 & 7.96 \\
     \checkmark & \checkmark & 53.3 & 36.5 & 74.0 & 99.9 & 7.02 \\
    \hline \hline
    \end{tabular}}
    \caption{Ablation studies of the context gates and auxiliary heads on MSVD benchmark. C.Gates denotes context gates, and A.Heads denotes auxiliary heads.}
    \label{tab:ablation_1}
\end{table}
The quantitative results in Table \ref{tab:main_table} show that our model performs significantly better than other methods, especially in CIDEr. CIDEr is specially designed for captioning tasks and is more similar to human judgment than other metrics. Compared to OA-BTG \cite{oa_btg:1} and ORG-TRL \cite{org_trl:1} our model lags in the BLEU-4 score, which is more suitable and designed for machine translation evaluation. Also, ORG-TRL \cite{org_trl:1} and OA-BTG \cite{oa_btg:1} have utilized object interaction and object trajectory features, respectively, whereas we have only taken mean localized object features for simplicity.
\subsection{Ablation Studies}
We perform the quantitative evaluation to investigate the effectiveness of the components we proposed. We conduct ablation studies that start with a primary decoder and add auxiliary heads and context gates incrementally. The Table \ref{tab:ablation_1} shows the results.

\begin{table}[h]
\centering
    \resizebox{0.8\columnwidth}{!}{
    \begin{tabular}{c |c c c }
    \hline \hline
        Caption &  \textit{OH} & \textit{AH} & COAHA    \\
    \hline
     A man is riding a motorcycle  & 0 & 0 & 0     \\
     A man is riding a \textbf{horse}  & 8.12 & 0 & 8.12     \\
      A man is riding a \textbf{car}  & 5.10 & 0 & 5.10     \\
      A \textbf{woman} is riding a motorcycle  & 4.67 & 0 & 4.67     \\
      An \textbf{animal} is riding a motorcycle  & 9.58 & 0 & 9.58     \\
      A man is \textbf{playing} with a motorcycle  & 0 & 9.52 & 9.52     \\
      A man is \textbf{eating} a motorcycle  & 0 & 10.4 & 10.4     \\
      A man is \textbf{playing} with a \textbf{toy}  & 6.45 & 9.52 & 15.97    \\

    \hline \hline
    \end{tabular}}
    \caption{Semantic significance of COAHA metric. $OH$ and $AH$ represent object and action hallucination scores respectively. For $OH$, $AH$ and COAHA, lower is better.}
    \label{tab:COAHA_significance}
\end{table}
\subsection{Validation of Context Gates}
\begin{figure}[t]
\begin{center}
    \includegraphics[width=0.7\textwidth]{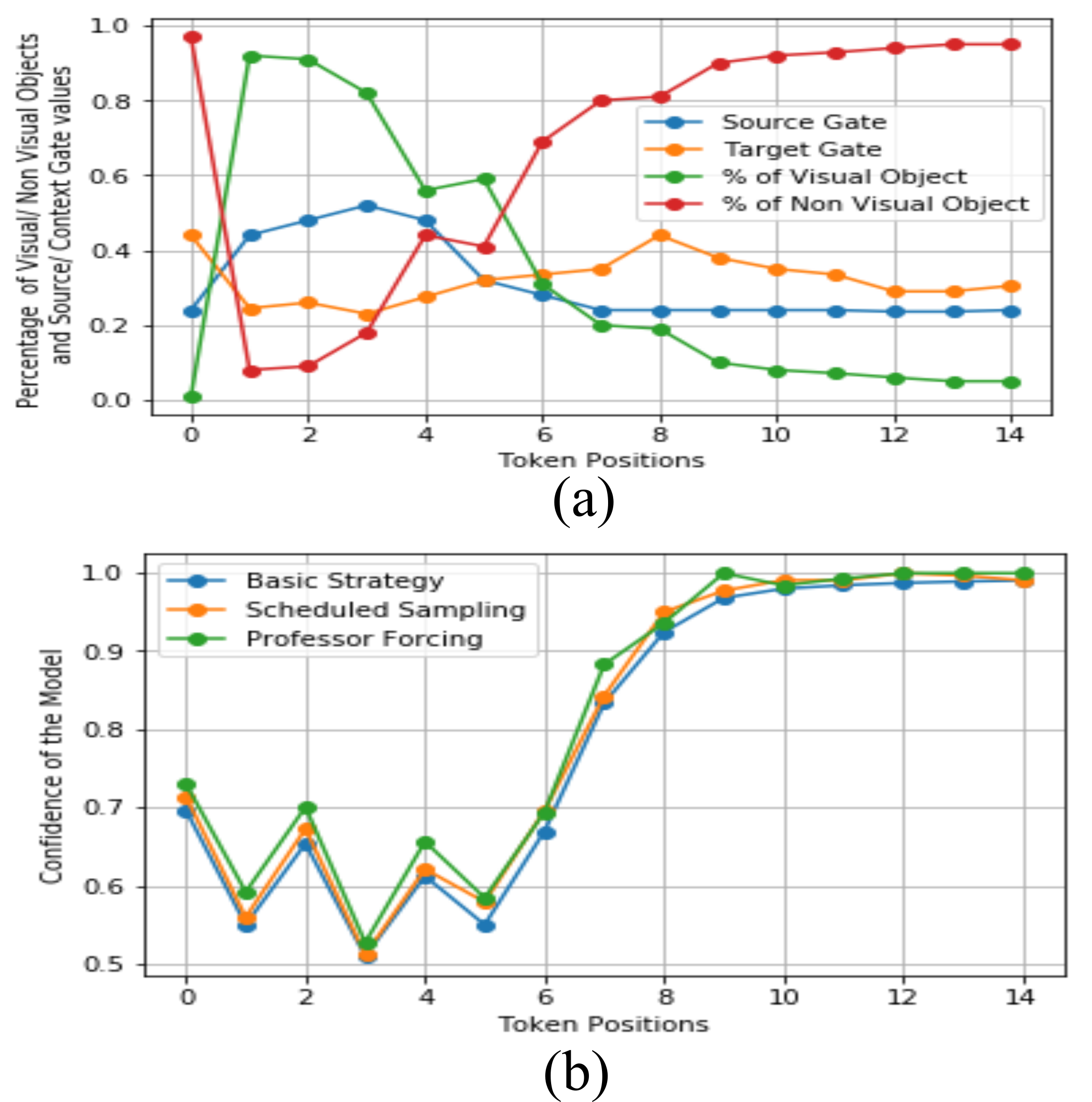}
    \caption{(a) Contribution of context gates during the generation of visual and non-visual words.
    (b) The model confidence vs. token positions under different training strategies.}
    \label{fig:context_gates_result}
    \end{center}
\end{figure}

To validate that the context gates are helping in dynamically selecting the feature importance, we have shown the values of context gates during multi-modal fusion. The Fig.~\ref{fig:context_gates_result}(a) shows that the source context gate value is high during the generation of visual object words. In contrast, the target context gate value is high during the generation of non-visual words.
\subsection{Effect of Exposure Bias on Hallucination }
In the case of machine translation, it is proven \cite{hal_nmt:1,hal_nmt:2} that exposure bias affects mainly under domain shift. In order to validate that exposure bias does not affect in same domain cases, we have shown the model confidence (Fig.~\ref{fig:context_gates_result}(b)) and COAHA values (Table \ref{tab:COAHA_table}) at different training strategies. Although the scheduled sampling \cite{schedule_sampling:1} and the professor forcing \cite{professor_forcing:1} are designed to mitigate exposure bias, they do not affect hallucination significantly.
\begin{table}[h]
\centering
    \resizebox{0.5\columnwidth}{!}{
    \begin{tabular}{c |  c  c}
    \hline \hline
        Training Strategy &  COAHA &   \\
    \hline
      Teacher Forcing  & 7.02 &   \\
      Scheduled Sampling & 7.01 &   \\
      Professor Forcing & 6.98 &   \\
    \hline \hline
    \end{tabular}}
    \caption{COAHA scores at different training strategies.}
    \label{tab:COAHA_table}
\end{table}
\subsection{Significance of COAHA} In order to show the semantic significance of our proposed metric, we have calculated OH (Object hallucination rate), AH (Action hallucination rate), and COAHA for different semantically perturbed captions, as shown in Table \ref{tab:COAHA_significance}.
\begin{figure}[t]
\begin{center}
    \includegraphics[width=0.75\textwidth]{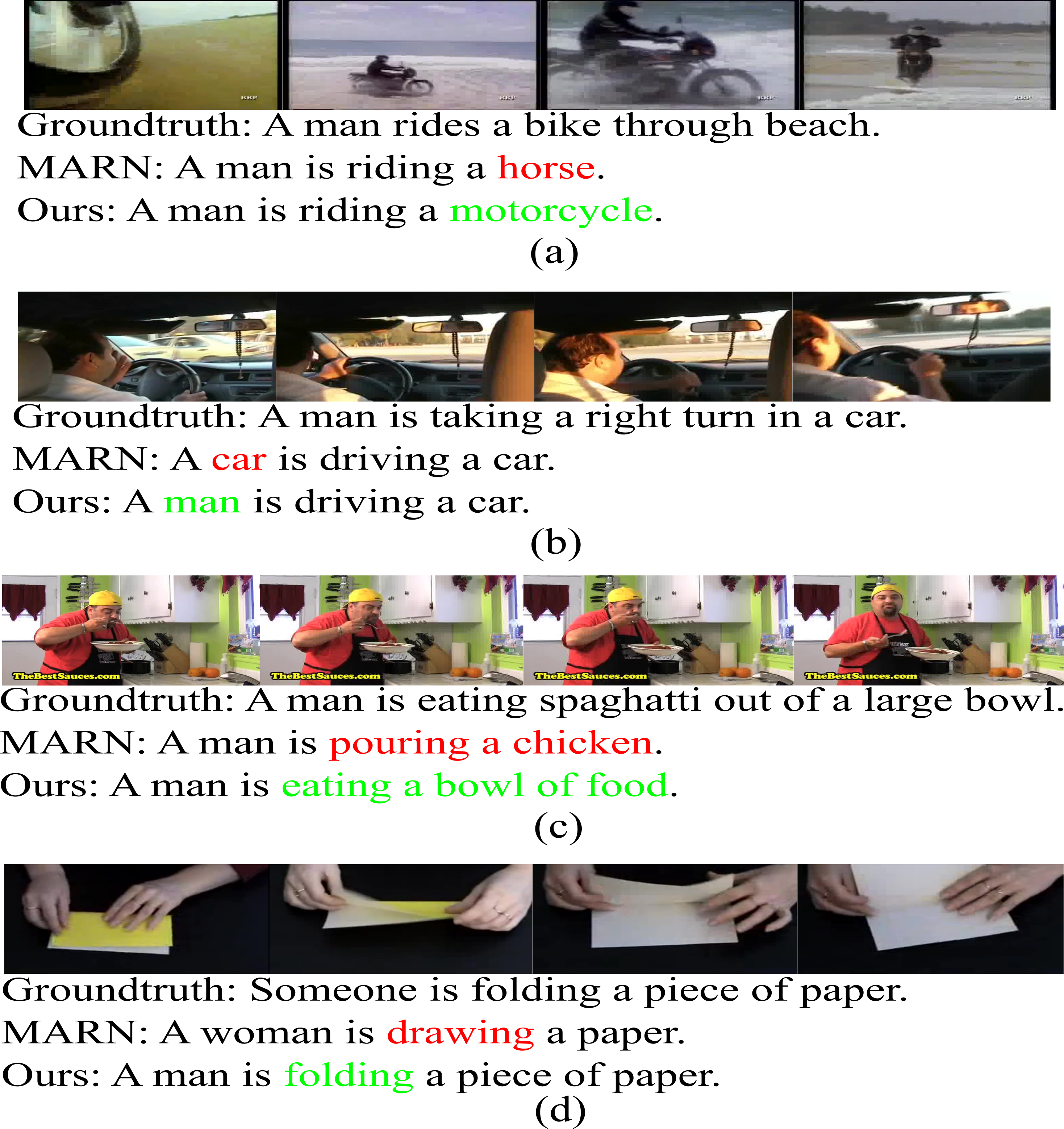}
    \caption{Captions generated by our model and MARN \cite{marn:1}.}
    \label{fig:qualitative_result}
    \end{center}
\end{figure}
\subsection{Qualitative Results}
We have compared the captions generated by our model and by Pei et al. \cite{marn:1}. From Fig.~\ref{fig:qualitative_result}, we can see that our model is less prone to hallucination and better identify objects and actions in the video.

\section{ Conclusion}
This work shows that hallucination is a significant problem for video captioning and points out three significant hallucination sources in the existing framework. We propose two robust and straightforward strategies to mitigate hallucination. We have got new state-of-the-art on two popular benchmark datasets by applying our solutions, especially by a significant margin in CIDEr score. Furthermore, we have shown that exposure bias is not a significant issue in same domain cases. Finally, we propose a new metric to measure the hallucination rate that the standard metrics cannot capture. Qualitative and quantitative experimental results show our model's effectiveness in overcoming the hallucination problem in video captioning.

\bibliographystyle{splncs04}
\bibliography{mybibliography}





\end{document}